\documentclass[10pt,twocolumn,letterpaper]{article}

\usepackage[pagenumbers]{cvpr} 

\usepackage{multirow}
\usepackage{graphicx}
\usepackage{amsmath}
\usepackage{amssymb}  
\usepackage{bbding}
\usepackage{subcaption}  
\usepackage{colortbl}  
\usepackage[svgnames]{xcolor}  

\definecolor{cvprblue}{rgb}{0.21,0.49,0.74}
\usepackage[pagebackref,breaklinks,colorlinks,allcolors=cvprblue]{hyperref}

\def\paperID{10142} 
\def\confName{CVPR}
\def\confYear{2026}

\title{Remedying Target-Domain Astigmatism for Cross-Domain Few-Shot \\Object Detection}

\author{
  Yongwei Jiang \quad
  Yixiong Zou\thanks{Corresponding author.} \quad
  Yuhua Li \quad 
  Ruixuan Li\\
  \textnormal{School of Computer Science and Technology, Huazhong University of Science and Technology}
  \\
  \texttt{\small \{jiangyongwei, yixiongz, idcliyuhua, rxli\}@hust.edu.cn}
  \\
}

\begin{document}
\maketitle

\begin{abstract}
Cross-domain few-shot object detection (CD-FSOD) aims to adapt pretrained detectors from a source domain to target domains with limited annotations, suffering from severe domain shifts and data scarcity problems. In this work, we find a previously overlooked phenomenon: models exhibit dispersed and unfocused attention in target domains, leading to imprecise localization and redundant predictions, just like a human cannot focus on visual objects. Therefore, we call it the target-domain \textbf{Astigmatism} problem. 
Analysis on attention distances across transformer layers reveals that regular fine-tuning inherently shows a trend to remedy this problem, but results are still far from satisfactory, which we aim to enhance in this paper.
Biologically inspired by the human fovea-style visual system, we enhance the fine-tuning's inherent trend through a center-periphery attention refinement framework, which contains (1) a Positive Pattern Refinement module to reshape attention toward semantic objects using class-specific prototypes, simulating the visual center region;
(2) a Negative Context Modulation module to enhance boundary discrimination by modeling background context, simulating the visual periphery region; and (3) a Textual Semantic Alignment module to strengthen center-periphery distinction through cross-modal cues. 
Our bio-inspired approach transforms astigmatic attention into focused patterns, substantially improving adaptation to target domains. Experiments on six challenging CD-FSOD benchmarks consistently demonstrate improved detection accuracy and establish new state-of-the-art results.
\end{abstract}    
\section{Introduction}
Great progress has been made in object detection~\cite{chen2017multi,tabernik2020segmentation,esteva2021deep}, thanks to the development of pretrained models on large-scale general data. However, real-world applications always require generalizing pretrained models to downstream expert domains, such as medical diagnosis~\cite{mti2030047,cmc} and industrial inspection~\cite{zheng2021recent}, where sufficient training samples are hard to collect, making it difficult to adapt the pretrained model. To address this issue, the cross-domain few-shot object detection (CD-FSOD)~\cite{fu2024cross} task has been proposed, which aims to pretrain an object detection model on a source-domain dataset and then adapt it to target domains with only scarce training data.
The domain gap and data scarcity have made it a challenging and unsolved problem.
\begin{figure}[t]
    \centering
    \includegraphics[width=\linewidth]{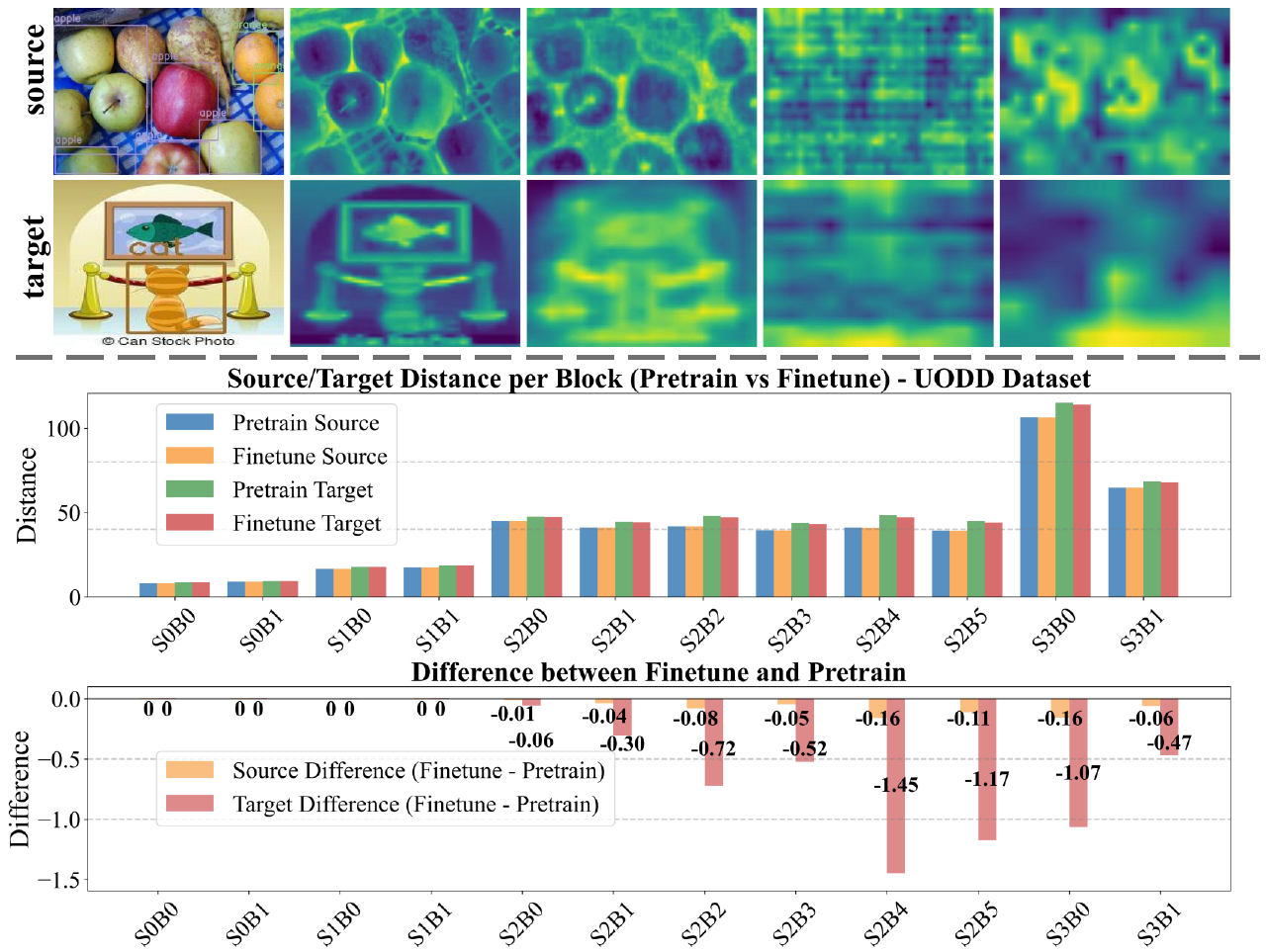}
    \caption{
    Visualization and quantification of the target-domain \textbf{Astigmatism} problem.
    \textbf{Top:} Attention maps measured across transformer blocks show that, in source domains (first row), attention progressively focuses on foreground objects, while in target domains (second row), attention remains persistently dispersed, resulting in oversized boxes and redundant predictions in object detection.
    \textbf{Bottom:} Attention distance across network depth 
   (\textit{SxBy} denotes Stage~x, Block~y in the Swin Transformer)
   reveals: 
    (1) a observed rise-then-fall trend in attention distance, reflecting an initial broad attention that gradually concentrates on objects for precise localization; 
    (2) consistently higher attention dispersion in target domains compared to source domains; 
    and (3) regular fine-tuning only marginally reduces this attention dispersion.}
    \label{fig:attention_distance_analysis}
\end{figure}


Existing CD-FSOD approaches have explored various techniques, including distillation-based methods~\cite{xiong2023cd} and domain‑adaptive optimization strategies~\cite{gao2023asyfod,fu2024cross}. However, the attention of detection models across different domains is rarely studied. To delve into it, we conduct an in-depth analysis of the model's attention across transformer blocks and discover an interesting phenomenon (Fig.~\ref{fig:attention_distance_analysis} top): the model experiences \textbf{Astigmatism} on target domains.
That is, in source domains, attention progressively concentrates on foreground objects as network depth increases, especially at the last two blocks.
By contrast, in target domains, attention remains dispersed and unfocused, leading to oversized bounding boxes and redundant predictions for object detection, just like a human not knowing what to focus on in a new domain and just coarsely looking at the image.

To quantify this phenomenon, we measure the attention distance~\cite{raghu2021vision} of each layer in the encoder as $\bar{d} = \frac{1}{N}\sum_{i=1}^{N} \sum_{j=1}^{N} A_{ij} \cdot ||p_i - p_j||$, where $A_{ij}$ represents attention weights between tokens, and $p_i$ and $p_j$ denote spatial positions, e.g., (0, 0) or (14, 14). 
As shown in Fig.~\ref{fig:attention_distance_analysis} bottom, we find that: 
(1) The attention distance follows a rise-then-fall trend with the network depth growing, indicating the learned patterns grow from local to global and finally return to focused ones for accurate localization.
(2) Target domains consistently show higher attention distance than source domains, indicating dispersed attention and ineffective feature extraction.
(3) Regular finetuning consistently tries to reduce this dispersion, but can only marginally achieve this goal, as the \textit{Target Differences} are all negative, but the \textit{Fintune Targets} are still higher than \textit{Finetune Sources} in Fig.~\ref{fig:attention_distance_analysis} bottom. 
These results validate that Astigmatism widely exists in CD-FSOD, and the model inherently tries to remedy this problem via regular finetuning. However, due to domain gaps and data scarcity, results are still far from satisfactory.

In this paper, we aim to enhance the model's inherent trend in remedying the Astigmatism problem, helping the model develop a concentrated attention on the semantic object for CD-FSOD.
To handle this human-like problem, we also take inspiration from human visual systems.
As shown in Fig.~\ref{fig:motivation}, humans possess a fovea-style visual system, where the center perception zone is dedicated to capturing highly detailed visual information, while the peripheral zones capture less details~\cite{laubrock2013control,weiss2014associating}. Such a center-peripheral visual property keeps human attention concentrated on the center zone~\cite{stewart2020review}.
Biologically inspired by this center-peripheral property of human visual systems, we design three complementary modules to enhance the representation of both the core and peripheral perception zones, and strengthen the contrast of these zones.
Specifically, we design
a Positive Pattern Refinement module that reshapes attention toward foreground objects by leveraging class-specific prototypes (central region); 
a Negative Context Modulation module that enhances object-background boundaries by explicitly modelling background contexts (peripheral region); 
and a Textual Semantic Alignment mechanism that enforces these distinctions through cross-modal knowledge integration using ``not [class]'' descriptors (e.g., ``not sofa, not dog'') to establish clearer foreground-background separation. This multi-faceted approach effectively transforms the astigmatic attention patterns in target domains into focused ones, improving target-domain performance. In summary:


\begin{figure}[t]
    \centering
    \includegraphics[width=0.45\textwidth]{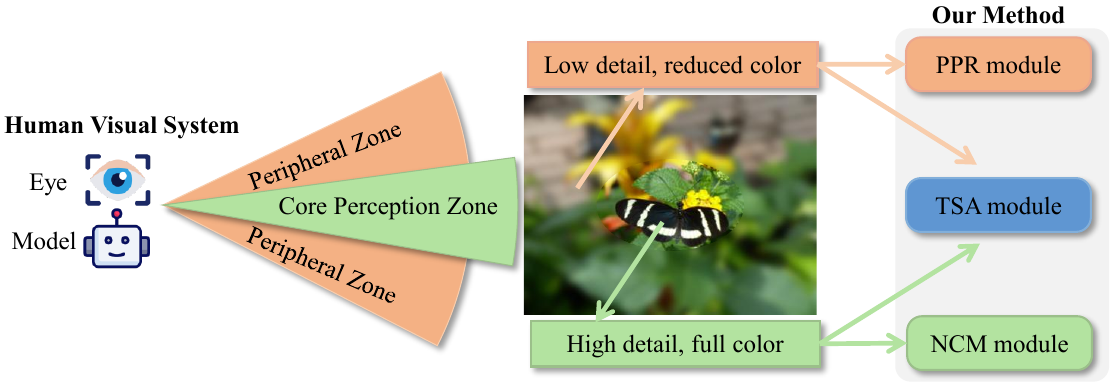}
    \caption{The inspiration from human fovea-style vision to remedy the Astigmatism problem. Our method mimics the human visual system: the Core Perception Zone (green) with high-detail processing guides the Positive Pattern Refinement (PPR) module to reshape attention more effectively toward foreground objects, while the Peripheral Zone (orange) with reduced details informs the Negative Context Modulation (NCM) module to enhance object-background boundaries by modeling background contexts. The Textual Semantic Alignment (TSA) module enforces the distinctions between center and peripheral regions, analogous to the center-surround mechanism of biological perception.}
    \label{fig:motivation}
\end{figure}


\begin{itemize}
    \item To the best of our knowledge, we are the first to find the target-domain Astigmatism problem, where the model exhibits dispersed attention that is harmful for object detection. Though the model tries to remedy it through the regular finetuning, the result is still far from satisfactory.
    
    \item Inspired by the human visual system, we propose a center-peripheral prototype-based method to enhance the model's inherent trend in remedying the Astigmatism problem, containing a Positive Pattern Refinement (PPR) module, a Negative Context Modulation (NCM) module, and a Textual Semantic Alignment (TSA) module.
    
    \item In the PPR module, we design to leverage class-specific prototypes to enhance the visual center region for the foreground attention. In the NCM module, we model background contexts for enhancing the visual peripheral region. In the TSA module, we strengthen the distinctions between center and peripheral regions through cross-modal knowledge integration.
    
    \item We demonstrate consistent performance improvements across six challenging cross-domain datasets, establishing new state-of-the-art results in the CD-FSOD task.
\end{itemize}

\begin{figure*}[t]
    \centering
    \includegraphics[width=0.99\textwidth]{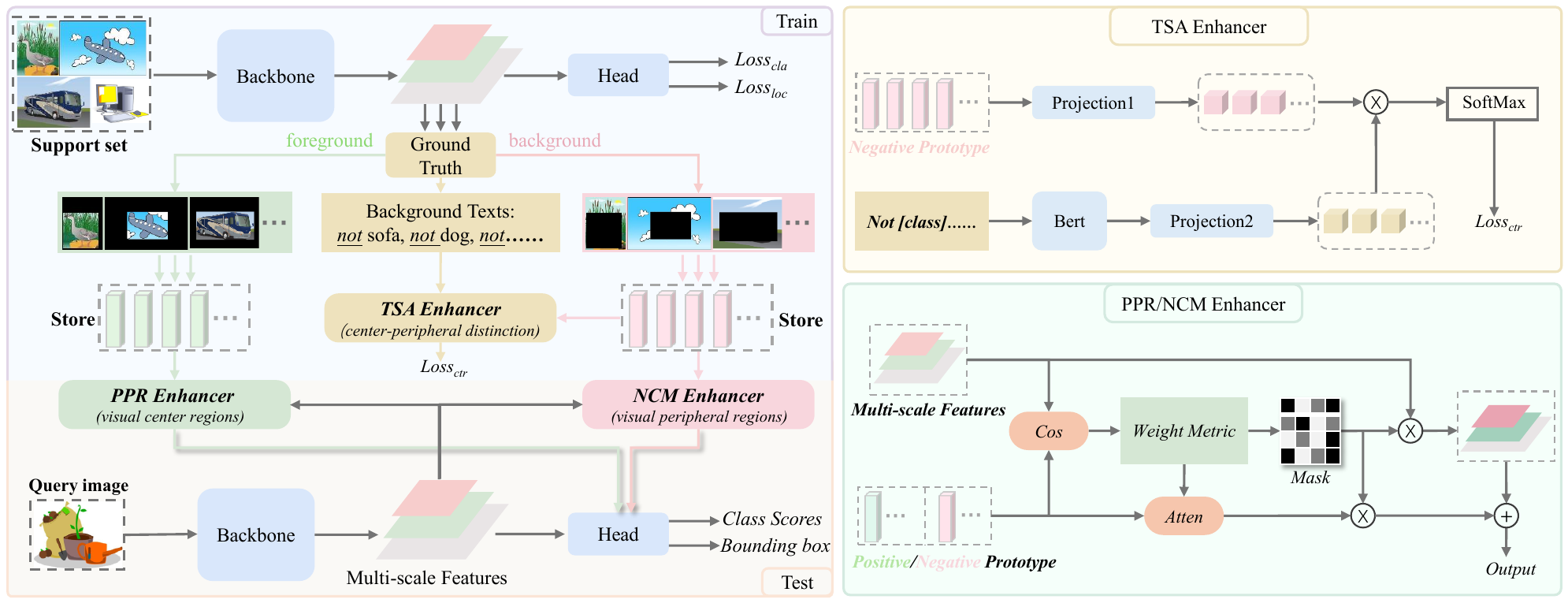}\vspace{-0.3cm}
    \caption{Overview of our human-vision-inspired framework to remedy the Astigmatism problem in CD-FSOD. The architecture integrates three complementary modules: (1) Positive Pattern Refinement (PPR) reshapes attention toward foreground objects using class prototypes; (2) Negative Context Modulation (NCM) enhances object-background boundaries through explicit background modeling; and (3) Textual Semantic Alignment (TSA) enhances these distinctions via cross-modal knowledge integration with negative descriptors (``not [class]''). During training (top), the model is optimized with both detection and alignment objectives, and extracts discriminative prototypes from support examples, stored in positive and negative repositories. At inference (bottom), stored prototypes turn dispersed attention patterns in query images into crystallized object-centric representations, analogous to the human center-peripheral visual system.}
    \label{fig:overview}
    \vspace{-3mm}
\end{figure*}

\section{Related Work}

\subsection{Cross-Domain Few-Shot Object Detection}
Cross-Domain Few-Shot Object Detection (CD-FSOD) seeks to effectively tackle the dual challenges of domain shift and data scarcity~\cite{zou2024closer,zou2024attention,zou2024compositional,zou2024flatten,zou2022margin,zhao2026interpretable,zhang2026mind}. Early works tackled these issues separately—FAFR-CNN~\cite{wang2019few} introduced domain-adaptive detection with limited examples via pairing, while PICA~\cite{zhong2022pica} explored instance-level feature alignment. DAPN~\cite{zhao2021domain} attempts to unify few-shot learning and domain adaptation within a joint framework. Related efforts such as AcroFOD~\cite{gao2022acrofod} applys augmentation in broader cross-domain settings. Additional methods like SDAFL~\cite{shi2020sensor} and FUM~\cite{yuan2022novel} explore domain-specific modalities and uncertainty modeling. Integrated approaches followed, with MoF-SOD~\cite{inoue2019cross} analyzing architecture effects on generalization, Distill-CDFSOD~\cite{xiong2023cd} using knowledge distillation to retain source priors, and CD-ViTO~\cite{fu2024cross} establishing benchmarks across diverse domains. 
Unlike prior work~\cite{jiang2025revisiting,zhang2024learning,zhang2024micm,zhang2025decoupling,zhang2026reclaiming}, we tackle the underexplored issue of cross-domain attention dispersion, improving knowledge transfer under diverse categories and visual shifts in few-shot settings.\vspace{-2mm}

\subsection{Context Modeling and Feature Transformation}
\vspace{-1mm}
Recent works have emphasized contextual information patterns as key for visual understanding tasks. Approaches like Dual Semantic Guidance~\cite{wang2025dual} use background screening modules to filter irrelevant visual information, while CDFormer~\cite{meng2024cdformer} tackles feature confusion via explicit object-background distinguishing mechanisms. For domain adaptation, IPNet~\cite{zhang2024ipnet} develops separate foreground and background domain alignment paths with specialized discriminators. Beyond detection, DualAnoDiff~\cite{jin2024dualanodiff} uses background cues for content consistency within surrounding contexts in anomaly detection, while NegativePrompt~\cite{wang2024negativeprompt} studies how negative contextual elements shape model performance in language domains. These methods overlook the distinctive cross-domain feature dispersion phenomenon that we reveal, and fail to exploit negative contextual cues for enhancing cross-modal representations.\vspace{-3mm}
\section{Method}
\vspace{-2mm}
In this section, we introduce our bio-inspired approach for remedying the Astigmatism problem in CD-FSOD. As illustrated in Figure~\ref{fig:overview}, our framework integrates Positive Pattern Refinement (PPR) to reshape attention toward foreground objects, working alongside Negative Context Modulation (NCM) which enhances object-background boundaries. Complementing these, Textual Semantic Alignment (TSA) enhances distinctions via cross-modal knowledge.
\subsection{Preliminaries}
\paragraph{Cross-Domain Few-Shot Object Detection (CD-FSOD)} adapts a model trained on a source domain with abundant labeled data (typically COCO~\cite{coco}) to a target domain $\mathcal{D}_T = \{(I_i^T, \mathcal{B}_i^T, \mathcal{C}_i^T)\}_{i=1}^{N_T}$ with sparse annotations and domain gap.
Here, $I_i$ represents images, $\mathcal{B}_i$ denotes bounding boxes, and $\mathcal{C}_i$ indicates class labels. 
The model is first trained on the source domain, then fine-tuned using a compact support set $\mathcal{S} = \{(I_i^T, \mathcal{B}_i^T, \mathcal{C}_i^T)\}_{i=1}^{N \times K}$ containing exactly $K$ annotated examples for each of the $N$ novel classes. Finally, the model is evaluated on previously unseen target domain images from the query set $\mathcal{Q}$.
In this paper, we solely focus on the fine-tuning stage in the target domain.

\vspace{-4mm}

\paragraph{Grounded Language-Image Pre-training (GLIP)} serves as our baseline model for object detection. Given an input image $I$, GLIP extracts visual features through a backbone: $\mathbf{F}_v = \{\mathbf{f}_v^1, \mathbf{f}_v^2, ..., \mathbf{f}_v^L\} = \text{Backbone}(I)$, where $\mathbf{F}_v$ represents the set of multi-scale visual features at $L$ different scales, and $\mathbf{f}_v^s$ denotes the feature map at scale $s$. For text inputs, GLIP employs BERT to process queries, yielding textual features $\mathbf{F}_t$. The model subsequently employs a fusion mechanism that combines these representations: $\mathbf{F}_{fused} = \text{DynamicConv}(\mathbf{F}_v, \mathbf{F}_t)$. These fused features are then fed into detection heads to produce class logits and corresponding bounding box predictions. The baseline detection loss is therefore formulated as:
\begin{equation}
\mathcal{L}_{detection} = \mathcal{L}_{cls} + \mathcal{L}_{loc}
\end{equation}
where $\mathcal{L}_{cls}$ and $\mathcal{L}_{loc}$ represent classification and localization losses, respectively.


\begin{figure}[t]
  \centering
  \includegraphics[width=0.7\linewidth]{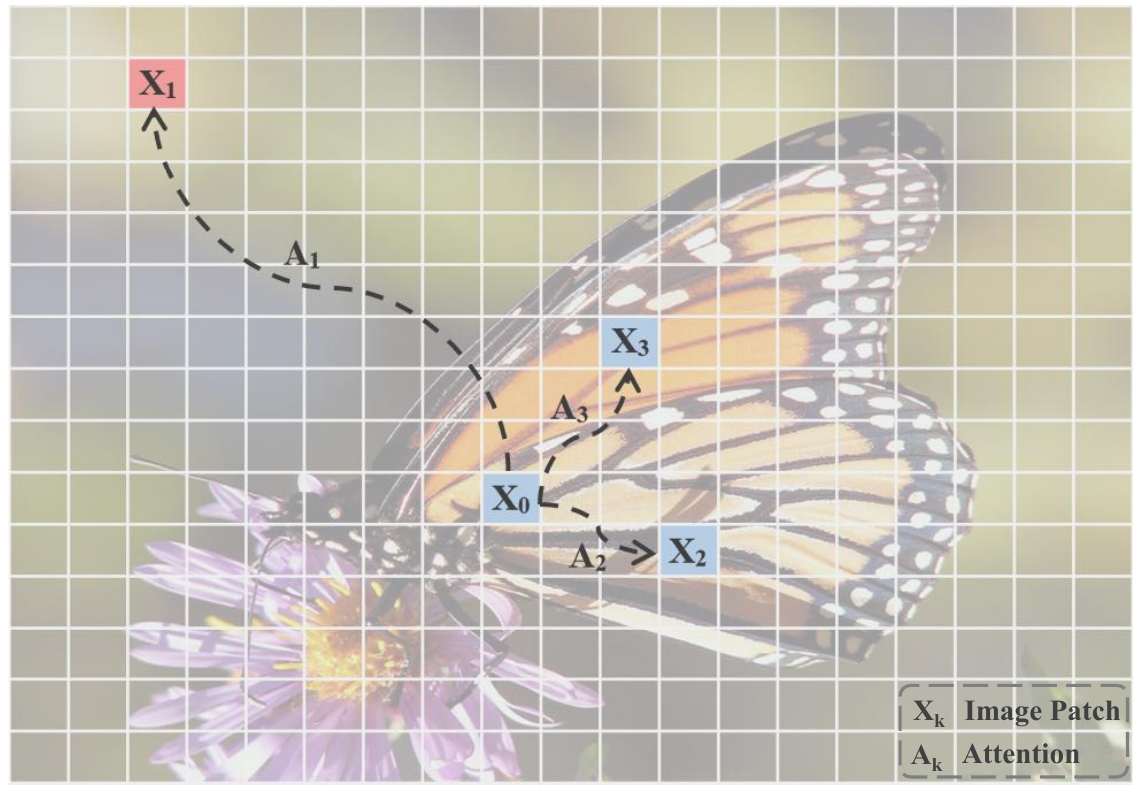}
  \vspace{-2mm}
  \caption{Attention distribution around a foreground patch $x_0$. Dashed arrows labeled $A_k$ indicate attention from $x_0$ to neighbors $x_k$. Same‑object neighbors $x_2,x_3$ are spatially near and should receive high attention, whereas the background patch $x_1$ is distant and should be weak. In target‑domain Astigmatism, domain shift diverts attention toward the distant background ($A_1\uparrow$; $A_2,A_3\downarrow$), yielding dispersed attention and a larger attention distance. Since the spatial distance between patches are fixed, our method reverses this dispersion by (i) down‑weighting background responses $A_1$ via a learned background prototype and simple ``not [class]'' cues (NCM/TSA module) to suppress spurious foreground–background affinity, and (ii) up‑weighting same‑object responses $A_2,A_3$ via class‑specific foreground prototypes (PPR module) to strengthen intra‑object compatibility, thereby shortening attention distance and restoring a focused, object‑centric pattern.}\vspace{-1mm}
  
  \label{fig:attention_redistribution}
  \vspace{-2mm}
\end{figure}
\vspace{-2mm}


\begin{table*}[t]
\caption{1-shot cross-domain detection performance comparison (mAP). $^\dagger$ indicates results reported in CD-ViTO~\cite{fu2024cross}. Methods with * are reimplemented by us using Swin-Tiny backbone. GLIP and our method are also implemented with Swin-Tiny backbone.}
\vspace{-2mm}
\centering
\footnotesize
\setlength{\tabcolsep}{4pt}
\begin{tabular}{l|c|c|c|c|c|c|c|c|c}
\hline
Method & Backbone & Venue & ArTaxOr & Clipart1k & DIOR & DeepFish & NEU-DET & UODD & Avg. \\
\hline
Distill$^\dagger$~\cite{10096216} & ResNet50 & ICASSP'23 & 5.10 & 7.60 & 10.50 & -- & -- & 5.90 & / \\
ViTDet$^\dagger$~\cite{li2022exploring} & ViT-B/14 & ECCV'22 & 5.90 & 6.10 & 12.90 & 0.90 & 2.40 & 4.00 & 5.37 \\
ViTDet*~\cite{li2022exploring} & Swin-T & ECCV'22 & 8.85 & 10.22 & 18.50 & 1.06 & 4.54 & 4.90 & 8.01 \\
Detic$^\dagger$~\cite{zhou2022detecting} & ViT-L/14 & ECCV'22 & 3.20 & 15.10 & 4.10 & 9.00 & 3.80 & 4.20 & 6.57 \\
Detic*~\cite{zhou2022detecting} & Swin-T & ECCV'22 & 4.45 & 19.22 & 5.70 & 10.49 & 7.15 & 5.12 & 8.69 \\
DE-ViT$^\dagger$~\cite{zhang2024detect} & ViT-L/14 & arXiv'23 & 10.50 & 13.00 & 14.70 & 19.30 & 0.60 & 2.40 & 10.08 \\
DE-ViT*~\cite{zhang2024detect} & Swin-T & arXiv'23 & 14.81 & 16.06 & 18.00 & 22.40 & 1.20 & 2.97 & 12.57 \\
GLIP~\cite{li2021grounded} & Swin-T & CVPR'22 & 21.40 & 33.35 & 19.09 & 27.23 & 8.84 & 3.99 & 18.98 \\
CD-ViTO$^\dagger$~\cite{fu2024cross} & ViT-L/14 & ECCV'24 & 21.00 & 17.70 & 17.80 & 20.30 & 3.60 & 3.10 & 13.92 \\
CD-ViTO*~\cite{fu2024cross} & Swin-T & ECCV'24 & 20.85 & 21.83 & 18.90 & 23.53 & 6.82 & 3.85 & 15.96 \\
Domain-RAG*~\cite{li2025domainragretrievalguidedcompositionalimage} & Swin-T & NIPS'25 & 21.47 & 22.53 & 18.97 & 24.67 & 7.53 & 3.92 & 16.52 \\
VFM-MoE*~\cite{liudon} & Swin-T & NIPS'25 & 23.38 & 25.67 & 19.03 & 25.45 & 8.12 & 3.87 & 17.59 \\
\hline
Ours & Swin-T & -- & \textbf{27.93} & \textbf{38.75} & \textbf{25.67} & \textbf{30.43} & \textbf{12.88} & \textbf{7.18} & \textbf{23.81} \\
\hline
\end{tabular}
\label{tab:sota_1shot}
\end{table*}

\begin{table*}[t]
\caption{5-shot cross-domain detection performance comparison (mAP). $^\dagger$ indicates results reported in CD-ViTO~\cite{fu2024cross}. Methods with * are reimplemented by us using Swin-Tiny backbone. GLIP and our method are also implemented with Swin-Tiny backbone.}
\vspace{-2mm}
\centering
\footnotesize
\setlength{\tabcolsep}{4pt}
\begin{tabular}{l|c|c|c|c|c|c|c|c|c}
\hline
Method & Backbone & Venue & ArTaxOr & Clipart1k & DIOR & DeepFish & NEU-DET & UODD & Avg. \\
\hline
Distill$^\dagger$~\cite{10096216} & ResNet50 & ICASSP'23 & 12.50 & 23.30 & 19.10 & 15.50 & 16.00 & 12.20 & 16.43 \\
ViTDet$^\dagger$~\cite{li2022exploring} & ViT-B/14 & ECCV'22 & 20.90 & 23.30 & 23.30 & 9.00 & 13.50 & 11.10 & 16.85 \\
ViTDet*~\cite{li2022exploring} & Swin-T & ECCV'22 & 21.05 & 23.45 & 23.45 & 11.03 & 19.96 & 15.07 & 19.00 \\
Detic$^\dagger$~\cite{zhou2022detecting} & ViT-L/14 & ECCV'22 & 8.70 & 20.20 & 12.10 & 14.30 & 14.10 & 10.40 & 13.30 \\
Detic*~\cite{zhou2022detecting} & Swin-T & ECCV'22 & 10.63 & 20.31 & 14.78 & 17.46 & 20.80 & 14.07 & 16.34 \\
DE-ViT$^\dagger$~\cite{zhang2024detect} & ViT-L/14 & arXiv'23 & 38.00 & 38.10 & 23.40 & 21.20 & 7.80 & 5.00 & 22.25 \\
DE-ViT*~\cite{zhang2024detect} & Swin-T & arXiv'23 & 38.15 & 38.27 & 23.52 & 25.89 & 11.50 & 6.79 & 24.02 \\
GLIP~\cite{li2021grounded} & Swin-T & CVPR'22 & 49.11 & 39.28 & 28.39 & 28.40 & 19.55 & 11.40 & 29.36 \\
CD-ViTO$^\dagger$~\cite{fu2024cross} & ViT-L/14 & ECCV'24 & 47.90 & 41.10 & 26.90 & 22.30 & 11.40 & 6.80 & 26.07 \\
CD-ViTO*~\cite{fu2024cross} & Swin-T & ECCV'24 & 48.12 & 40.28 & 27.05 & 28.26 & 16.84 & 9.24 & 28.30 \\
Domain-RAG*~\cite{li2025domainragretrievalguidedcompositionalimage} & Swin-T & NIPS'25 & 46.13 & 41.47 & 25.23 & 27.67 & 17.53 & 9.87 & 27.98 \\
VFM-MoE*~\cite{liudon} & Swin-T & NIPS'25 & 48.23 & 39.13 & 26.34 & 27.58 & 18.67 & 10.73 & 28.45 \\
\hline
Ours & Swin-T & -- & \textbf{54.98} & \textbf{44.83} & \textbf{29.41} & \textbf{33.87} & \textbf{23.64} & \textbf{15.66} & \textbf{33.73} \\
\hline
\end{tabular}
\label{tab:sota_5shot}

\vspace{-3mm}
\end{table*}

\begin{table*}[t]
\caption{10-shot cross-domain detection performance comparison (mAP). $^\dagger$ indicates results reported in CD-ViTO~\cite{fu2024cross}. Methods with * are reimplemented by us using Swin-Tiny backbone. GLIP and our method are also implemented with Swin-Tiny backbone.}
\vspace{-2mm}
\centering
\footnotesize
\setlength{\tabcolsep}{4pt}
\begin{tabular}{l|c|c|c|c|c|c|c|c|c}
\hline
Method & Backbone & Venue & ArTaxOr & Clipart1k & DIOR & DeepFish & NEU-DET & UODD & Avg. \\
\hline
Distill$^\dagger$~\cite{10096216} & ResNet50 & ICASSP'23 & 18.10 & 27.30 & 26.50 & 15.50 & 21.10 & 14.50 & 20.50 \\
ViTDet$^\dagger$~\cite{li2022exploring} & ViT-B/14 & ECCV'22 & 23.40 & 25.60 & 29.40 & 6.50 & 15.80 & 15.60 & 19.38 \\
ViTDet*~\cite{li2022exploring} & Swin-T & ECCV'22 & 24.20 & 25.93 & 30.10 & 8.04 & 22.52 & 19.55 & 21.72 \\
Detic$^\dagger$~\cite{zhou2022detecting} & ViT-L/14 & ECCV'22 & 12.00 & 22.30 & 15.40 & 17.90 & 16.80 & 14.40 & 16.47 \\
Detic*~\cite{zhou2022detecting} & Swin-T & ECCV'22 & 13.80 & 22.59 & 17.20 & 22.04 & 23.97 & 16.58 & 19.36 \\
DE-ViT$^\dagger$~\cite{zhang2024detect} & ViT-L/14 & arXiv'23 & 49.20 & 40.80 & 25.60 & 21.30 & 8.80 & 5.40 & 25.18 \\
DE-ViT*~\cite{zhang2024detect} & Swin-T & arXiv'23 & 50.50 & 41.28 & 26.80 & 26.23 & 12.60 & 13.72 & 28.52 \\
GLIP~\cite{li2021grounded} & Swin-T & CVPR'22 & 62.01 & 41.04 & 31.10 & 29.29 & 20.56 & 19.56 & 33.93 \\
CD-ViTO$^\dagger$~\cite{fu2024cross} & ViT-L/14 & ECCV'24 & 60.50 & 44.30 & 30.80 & 22.30 & 12.80 & 7.00 & 29.62 \\
CD-ViTO*~\cite{fu2024cross} & Swin-T & ECCV'24 & 62.30 & 43.83 & 31.15 & 27.46 & 18.26 & 17.84 & 33.47 \\
Domain-RAG*~\cite{li2025domainragretrievalguidedcompositionalimage} & Swin-T & NIPS'25 & 59.87 & 42.85 & 29.23 & 28.67 & 18.87 & 17.93 & 32.81 \\
VFM-MoE*~\cite{liudon} & Swin-T & NIPS'25 & 63.13 & 41.23 & 30.67 & 29.13 & 20.23 & 19.23 & 33.94 \\
\hline
Ours & Swin-T & -- & \textbf{68.40} & \textbf{47.63} & \textbf{34.25} & \textbf{35.58} & \textbf{24.96} & \textbf{24.18} & \textbf{39.17} \\
\hline
\end{tabular}
\label{tab:sota_10shot}

\vspace{-4mm}
\end{table*}

\subsection{How to reduce attention distance to avoid Astigmatism?}
\vspace{-1mm}
In \cref{fig:attention_distance_analysis}, we validate the Astigmatism problem by measuring the attention distance. In this section, we aim to reduce such distances to avoid Astigmatism. Take a foreground patch $x_0$ as an example (Fig.~\ref{fig:attention_redistribution}), suppose it attends to $\{x_k\}$, let $A_k$ denote the attention from $x_0$ to $x_k$, $p_k$ is their spatial positions, and $r_k=\|p_0-p_k\|$ is the spatial distances, the attention distance is calculated as\vspace{-1mm}
\begin{equation}
d(x_0) = \sum_k A_k\,r_k.
\label{eq:attention-distance}
\end{equation}
For illustration \emph{(without loss of generality)}, consider one background neighbor $x_1$ and two same-object neighbors $x_2,x_3$ (omitting heads, batch indices, etc.), which leads to\vspace{-2mm}
\begin{equation}
d(x_0) = A_1 r_1 + A_2 r_2 + A_3 r_3.
\label{eq:d-three-neighbors}
\end{equation}
In target domains, we consistently observe larger attention distances (Fig.~\ref{fig:attention_distance_analysis}). Domain shift induces \emph{foreground–background confusion}: the distant background neighbor $x_1$ collects more attention ($A_1\uparrow$) while the attention to same‑object neighbors $x_2,x_3$ shrinks ($A_2,A_3\downarrow$). With fixed spatial offsets $r_k$ and $r_1>r_2,r_3$, this attention distribution necessarily increases $d(x_0)$, manifesting dispersed, target‑domain attention (Astigmatism). 

Therefore, to avoid Astigmatism, we need to rectify the attention distribution so that patches within the same object possess higher attention.
Our modules reallocate $\{A_k\}$ in a complementary way: (i) \textbf{PPR} uses class‑specific foreground prototypes to strengthen same‑object similarity and continuity along the object extent, increasing $A_2,A_3$; (ii) \textbf{NCM} builds a unified background prototype to enhance the representation of background, reducing its confusion with foreground and decreasing $A_1$; and (iii) \textbf{TSA} employs simple prompts ``not [class]'' to sharpen foreground–background separation via cross‑modal alignment, further decreasing $A_1$ and stabilizing gains on $A_2,A_3$. 
\vspace{-1mm}

\begin{table}[t]
\caption{Summary of target datasets used in our evaluation.}
\vspace{-2mm}
\centering
\footnotesize
\setlength{\tabcolsep}{2pt}
\begin{tabular}{lccccc}
\hline
\textbf{Dataset} & \textbf{Classes} & \textbf{Images} & \textbf{Boxes} & \textbf{Domain} & \textbf{Challenge} \\
\hline
ArTaxOr & 7 & 1383 & 1628 & photorealistic & small objects \\
Clipart1k & 20 & 500 & 1526 & cartoon & style abstraction \\
DIOR & 20 & 5000 & 28810 & aerial & medium-scale \\
DeepFish & 1 & 909 & 3029 & aquatic & poor visibility \\
NEU-DET & 6 & 360 & 834 & industrial & subtle features \\
UODD & 3 & 506 & 3218 & underwater & complex clutter \\
\hline
\end{tabular}
\label{tab:datasets}
\vspace{-2mm}
\end{table}
\vspace{-1mm}
\subsection{Positive Pattern Refinement}
\vspace{-1mm}
Based on the above analysis, we first design a Positive Pattern Refinement (PPR) module that uses class-specific prototypes from support examples to enhance foreground features, simulating the center region of human vision.

For clarity and simplicity, we present our method using single-scale notation, while our framework processes multi-scale features as shown in Figure \ref{fig:overview}.
During the finetuning phase, foreground prototypes are computed as:\vspace{-2mm}
\begin{equation}
\mathbf{p}_{fg}^{c} = \frac{1}{|\mathcal{B}_{fg}^{c}|} \sum_{(x,y) \in \mathcal{B}_{fg}^{c}} \mathbf{f}_v(x,y)
\end{equation}
where $\mathcal{B}_{fg}^{c}$ denotes the set of pixel coordinates within foreground regions of class $c$. 
Each prototype $\mathbf{p}_{fg}^{c} \in \mathbb{R}^{1 \times 1 \times D}$ represents the averaged feature representation for class $c$, where $D$ is the feature dimension. These prototypes are stored in $\mathcal{P}_{fg}$ to capture target object characteristics.

During inference, we first compute cosine similarity between each feature map position and stored prototypes:\vspace{-2mm}
\begin{equation}
\text{sim}(\mathbf{f}_v(x,y), \mathbf{p}_{fg}^{c}) = \frac{\mathbf{f}_v(x,y) \cdot \mathbf{p}_{fg}^{c}}{||\mathbf{f}_v(x,y)||_2 \cdot ||\mathbf{p}_{fg}^{c}||_2 + \epsilon}
\end{equation}

where $\mathbf{f}_v(x,y) \in \mathbb{R}^{1 \times 1 \times D}$ represents the feature vector at position $(x,y)$ in the feature map, $||\cdot||_2$ denotes the L2 norm, and $\epsilon = 10^{-8}$ ensures numerical stability.

We compute metrics via temperature-scaled softmax: $w_c(x,y) = \text{softmax}(\text{sim}(\mathbf{f}_v(x,y), \mathbf{p}_{fg}^{c})/T)$.
In parallel, we use a binary mask to flag high-similarity prototypes.
\begin{equation}
\mathbf{M}_{fg}(x,y) =
\begin{cases}
1, & \text{if } \max_c \text{sim}(\mathbf{f}_v(x,y), \mathbf{p}_{fg}^{c}) > \tau_{fg} \\
0, & \text{otherwise}
\end{cases}
\end{equation}
where $\tau_{fg}$ serves as the similarity threshold.
With this mask and weight metrics, we apply targeted feature enhancement:\vspace{-4mm}

\begin{align}
\mathbf{f}_v^{\text{pos}}(x,y) &= \mathbf{f}_v(x,y) \cdot \mathbf{M}_{fg}(x,y) \nonumber \\
&\quad + \gamma_{fg} \sum_c w_c(x,y) \mathbf{p}_{fg}^{c} \cdot \mathbf{M}_{fg}(x,y)
\end{align}
where $\gamma_{fg}$ controls the strength of prototype contribution. This selective approach ensures enhancement occurs precisely where objects of interest are likely to be present, effectively reshaping dispersed attention patterns into focused object-centric representations in target domains.
\vspace{-2mm}
\subsection{Negative Context Modulation}
\vspace{-1mm}
While foreground refinement is crucial, effective object detection equally depends on proper background modeling. Our NCM module complements the foreground focus by modeling contextual elements outside object regions, mimicking the peripheral region of human visual systems.

During the finetuning phase, we construct a unified background prototype from regions outside ground truth boxes:
\begin{equation}
\mathbf{p}_{bg} = \frac{1}{|\mathcal{B}_{bg}|} \sum_{(x,y) \in \mathcal{B}_{bg}} \mathbf{f}_v(x,y)
\end{equation}
where background regions $\mathcal{B}_{bg}$ are identified as:
\begin{equation}
\mathcal{B}_{bg} = \{(x,y) | (x,y) \notin \bigcup_{b \in \mathcal{B}_{gt}} b\}
\end{equation}
with $\mathcal{B}_{gt}$ denoting all ground truth boxes in the support set.

During the inference phase, following the same pipeline as PPR, we apply background-aware feature modulation:
\begin{equation}
\mathbf{f}_v^{\text{neg}}(x,y) = \mathbf{f}_v(x,y) \cdot \mathbf{M}_{bg}(x,y) + \gamma_{bg} \mathbf{p}_{bg} \cdot \mathbf{M}_{bg}(x,y)
\end{equation}
where $\mathbf{M}_{bg}(x,y)$ is computed using the same similarity-based thresholding as in PPR, and $\gamma_{bg}$ controls background contribution. 
Unlike the class-specific strategy in PPR, background is treated as a unified concept, enabling better separation of foreground and context in novel domains.
\vspace{-1mm}
\subsection{Textual Semantic Alignment}
Beyond visual feature refinement, leveraging semantic knowledge can further enhance domain adaptation. Our Textual Semantic Alignment (TSA) module enhances object-background distinctions via cross-modal knowledge. 

For each target domain, we encode negative class concepts using ``not [class]'' descriptors and utilize the background prototype from NCM as the visual representation:
\begin{equation}
\mathbf{F}_t^{bg} = \text{BERT}(\mathcal{T}_{bg}), \quad \mathbf{F}_v^{bg} = \mathbf{p}_{bg}
\end{equation}
where $\mathcal{T}_{bg}$ represents text prompts like ``not sofa, not dog'' corresponding to target classes. Importantly, all background texts employ \textbf{simple prompts} derived from dataset categories, which we describe in detail in the Appendix.


These visual and textual representations are aligned through projection into a shared semantic space:
\begin{equation}
\mathcal{S} = \mathbf{Proj}_1(\mathbf{F}_v^{bg}) \cdot \mathbf{Proj}_2(\mathbf{F}_t^{bg})^T
\end{equation}
where $\mathbf{Proj}_1$ and $\mathbf{Proj}_2$ are learnable projection layers, and $\cdot$ denotes matrix multiplication.

We optimize the alignment to maximize similarity between visual and textual background features:
\begin{equation}
\mathcal{L}_{ctr} = -\log\frac{\exp(\text{diag}(\mathcal{S})/\tau)}{\sum_{i,j} \exp(\mathcal{S}_{i,j}/\tau)}
\end{equation}
where $\tau$ is a temperature parameter. This formulation maximizes the diagonal elements (positive pairs) while minimizing off-diagonal ones (negative pairs). 
This cross-modal alignment offers extra supervision to remedy attention dispersion by creating clearer object-background boundaries.

\subsection{Model Training and Testing}
Our framework operates differently during finetuning and inference phases, as illustrated in Figure \ref{fig:overview}. 
The integration of our three complementary modules follows a coordinated architecture that remedies the Astigmatism problem through targeted feature transformation.

\textbf{During finetuning}, we first extract class-specific foreground prototypes and unified background prototypes from support examples. The model is then optimized with both detection and cross-modal alignment objectives:
\vspace{-2mm}
\begin{equation}
\mathcal{L}_{total} = \mathcal{L}_{detection} + \lambda_{bg} \cdot \mathcal{L}_{ctr}
\end{equation}
where $\lambda_{bg}$ balances detection and alignment objectives.

\textbf{During inference}, we apply the prototype-based enhancement through both PPR and NCM enhancers. To avoid potential conflicts between positive and negative enhancements, we use a complementary fusion strategy:
\vspace{-2mm}
\begin{equation}
\mathbf{F}_v^{\text{enhanced}} = \mathbf{f}_v^{\text{pos}} + \mathbf{f}_v^{\text{neg}}
\end{equation}



where $\mathbf{f}_v^{\text{pos}}$ and $\mathbf{f}_v^{\text{neg}}$ denote positive and negative enhanced features derived from similarities, enabling flexible regional enhancement.
The enhanced features $\mathbf{F}_v^{\text{enhanced}}$ are then fed into the baseline detection heads for class and box prediction.
This dual-pathway design enhances foreground features while preserving object–background boundaries, turning dispersed attention into focused representations and alleviating the astigmatism problem in CD-FSOD.\vspace{-2mm}

\section{Experiments}


\subsection{Dataset and Metrics}
\vspace{-1mm}


Our evaluation spans six cross-domain datasets (Table~\ref{tab:datasets}): ArTaxOr~\cite{ArTaxOr} (photorealistic arthropods, small objects), Clipart1k~\cite{Clipart1k} (cartoon abstractions), DIOR~\cite{DIOR} (aerial, medium-scale), DeepFish~\cite{FISH} (underwater visibility), NEU-DET~\cite{NEUDET} (industrial defects, imbalance), and UODD~\cite{UODD} (underwater organisms, complex, imbalanced). We report COCO-style mAP over IoU 0.5–0.95.
\vspace{-2mm}


\subsection{Benchmark Comparisons}
\vspace{-2mm}

Tables~\ref{tab:sota_1shot}, \ref{tab:sota_5shot}, and \ref{tab:sota_10shot} show our method consistently surpasses prior work in mAP across all few-shot settings. For fairness, methods marked * are reimplemented with Swin-Tiny. We deliver strong gains on every dataset and shot count, especially Clipart1k and DeepFish, where domain shifts are pronounced, validating that center-periphery attention refinement mitigates cross-domain attention dispersion.

\vspace{-2mm}
\subsection{Ablation Study}
\vspace{-1mm}
\begin{table}[t]
\caption{Ablation study showing each component's contribution to overall performance (mAP \%) in 5-shot setting.}
\vspace{-2mm}
\centering
\scriptsize
\setlength{\tabcolsep}{2pt}
\renewcommand{\arraystretch}{1.1}
\begin{tabular}{ccc|cccccc}
\hline
\multicolumn{3}{c|}{Components} & \multicolumn{6}{c}{Datasets} \\
\cline{1-3}\cline{4-9}
NCM & PPR & TSA & ArTaxOr & Clipart1k & DIOR & DeepFish & NEU-DET & UODD \\
\hline
 &  &  & 48.11 & 39.28 & 27.39 & 28.40 & 19.55 & 11.40 \\
\Checkmark &  &  & 49.95 & 40.52 & 27.75 & 29.58 & 20.48 & 12.21 \\
\Checkmark & \Checkmark &  & 52.68 & 42.95 & 28.62 & 31.96 & 22.38 & 14.26 \\
\Checkmark & \Checkmark & \Checkmark & \textbf{54.98} & \textbf{44.83} & \textbf{29.41} & \textbf{33.87} & \textbf{23.64} & \textbf{15.66} \\
\hline
\end{tabular}
\label{tab:ablation_study}
\vspace{-1mm}
\end{table}




Table \ref{tab:ablation_study} shows that NCM, PPR, and TSA yield average gains of +1.06\%, +2.06\%, and +1.59\%, respectively, with PPR showing the best improvement on DeepFish, confirming the need for balanced foreground–background modeling.

\begin{figure}[t]
\centering
\includegraphics[width=0.47\textwidth]{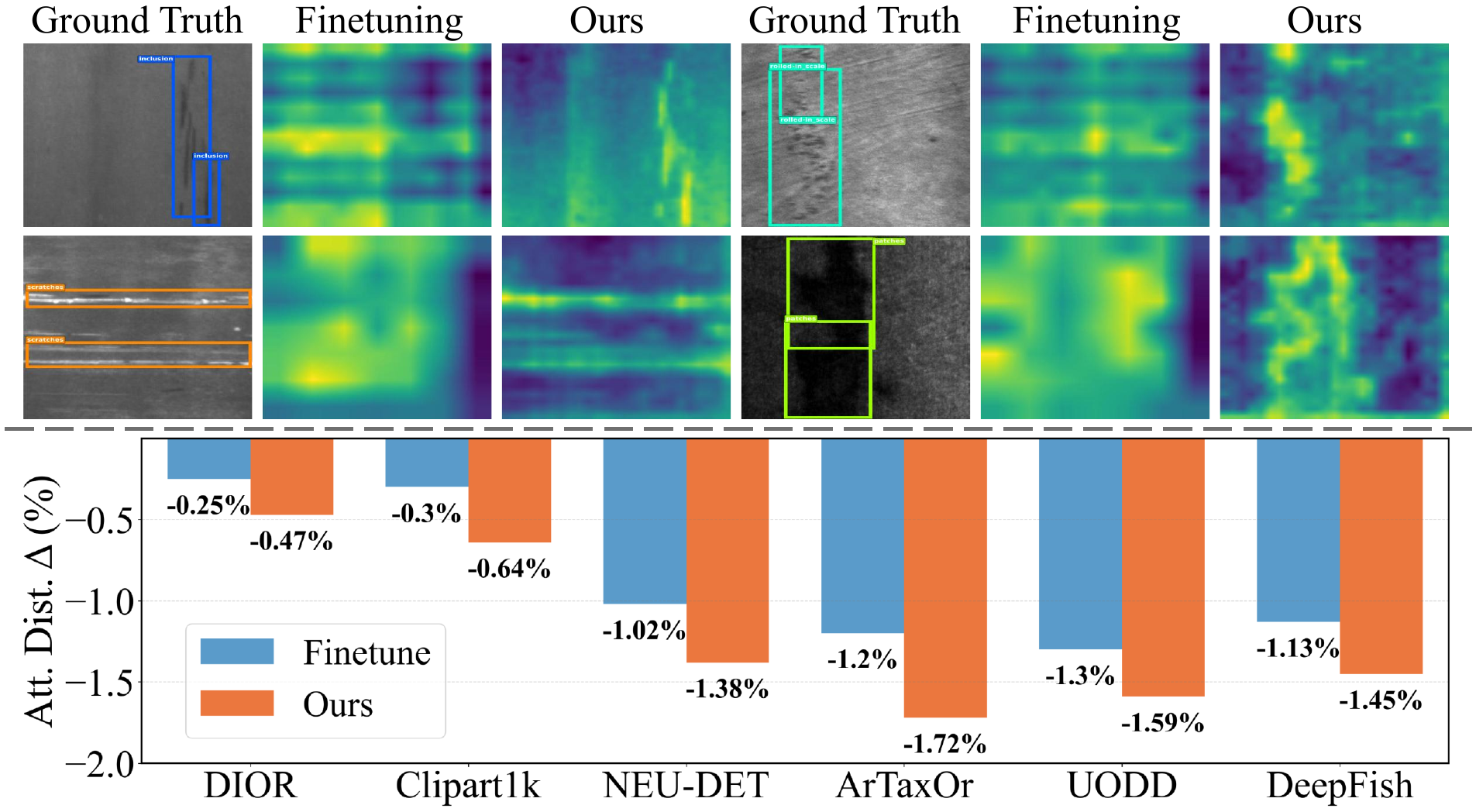}
\vspace{-2mm}
\caption{\textbf{Top:} Visualization of attention maps demonstrating the Astigmatism problem and our solution. Each row displays two different samples: including the original image with ground truth (left), conventional fine-tuning that only marginally alleviates dispersed attention patterns (middle), and ours, which indicates focused attention that precisely concentrates on target regions. \textbf{Bottom:} Change in attention distance relative to pretrained model across datasets. Negative values indicate reduction in attention dispersion (larger magnitude means better focus). While fine-tuning reduces dispersion, our method achieves substantially greater reductions, validating our superior ability to address the Astigmatism problem by reshaping attention toward foreground objects.}

\label{fig:attention_analysis}
\vspace{-6mm}
\end{figure}

\subsection{Visualization and Analysis}
\subsubsection{Analysis of Attention Pattern}

Figure~\ref{fig:attention_analysis} provides both visual and quantitative evidence of the Attentional Astigmatism problem and our solution's effectiveness. \textbf{Qualitatively}, the top panel shows conventional fine-tuning only marginally alleviates dispersed attention patterns, while our method yields significantly more focused attention that aligns closely with ground truth regions, even for subtle defects in the bottom row examples. 
\textbf{Quantitatively}, the bottom panel shows changes in attention distance relative to the pretrained baseline (negative indicate reduction). While conventional fine-tuning achieves modest reductions (0.25\%-1.30\%), our approach consistently achieves larger reductions (0.47\%-1.72\%) across all datasets, with the most significant improvements on challenging domains like ArTaxOr and DeepFish, validating our superior ability to address the Astigmatism problem.
\vspace{-1mm}

\begin{figure}[htp]
\vspace{-3mm}

\centering
\includegraphics[width=0.8\linewidth]{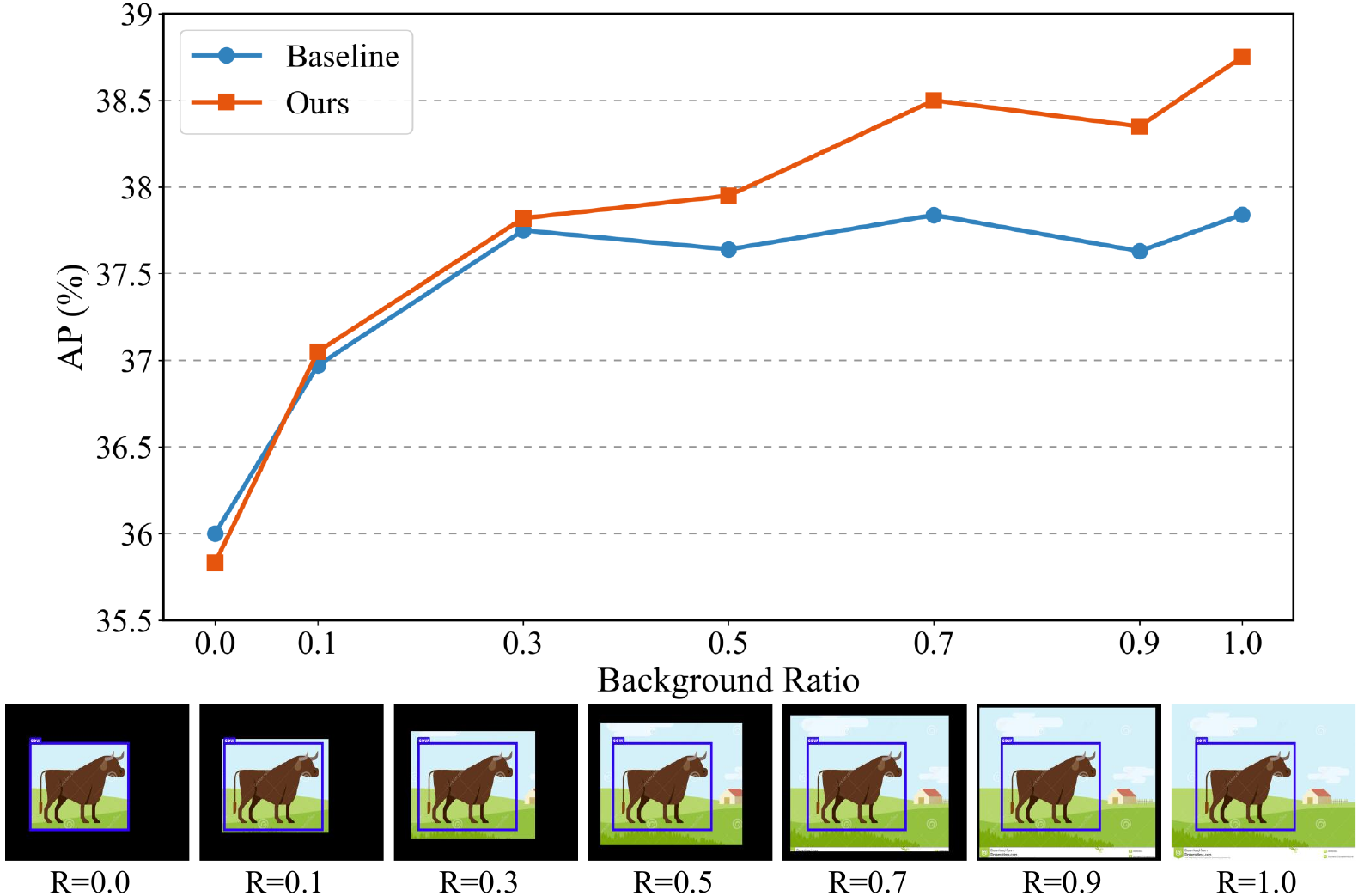}
\vspace{-2mm}
\caption{Validation of Negative Context Modulation (NCM) module through background ratio analysis. Top: Performance comparison with varying background ratios. Our method demonstrates consistent advantages over the baseline, particularly at higher background ratios (0.3-1.0), indicating NCM exploits negative background contexts as \textit{complementary cues to strengthen foreground feature learning and boundary discrimination}, thereby mitigating the attention dispersion in Astigmatism problem. Bottom: Visual examples showing detection with increasing background ratios from left to right (R=0.0 to R=1.0).}
\label{fig:background_ratio_analysis}
\vspace{-4mm}
\end{figure}
\subsubsection{Analysis of Background Context Utilization}

Figure~\ref{fig:background_ratio_analysis} validates the effectiveness of our Negative Context Modulation (NCM) module in addressing the Astigmatism problem. Both models show comparable performance at low background ratios, but our approach achieves significant gains at higher ratios (0.7-1.0). This performance pattern demonstrates that NCM successfully leverages background contextual information to enhance object-background boundary discrimination. By better utilizing background cues, the model implicitly improves foreground feature learning and localization, thereby easing the attention dispersion issue in cross-domain few-shot detection.\vspace{-1mm}

\subsubsection{Qualitative Detection Results}
\begin{figure}[htp]
    \centering
    \includegraphics[width=0.47\textwidth]{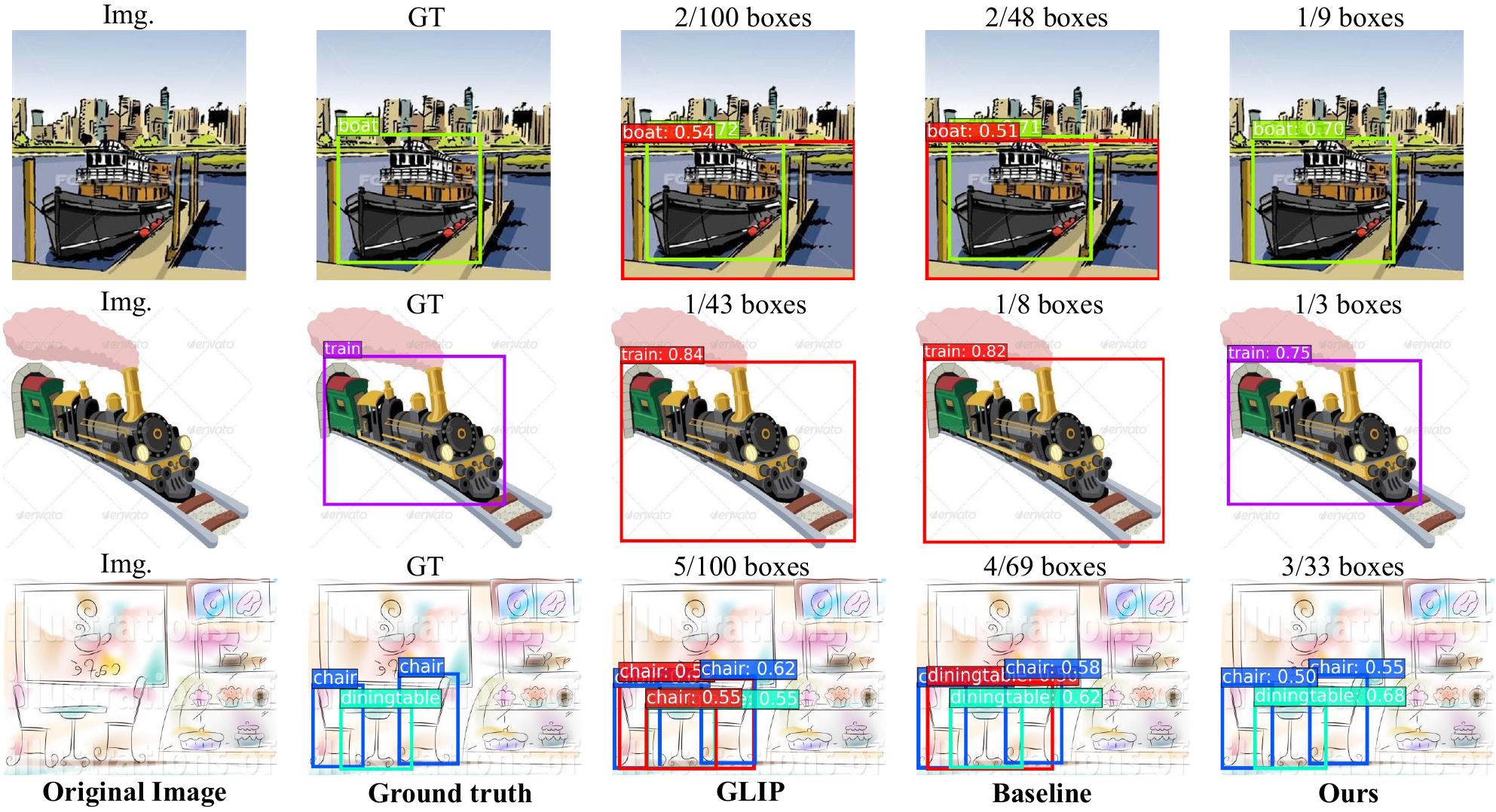}
    \vspace{-2mm}
    \caption{Qualitative detection results on cross-domain scenarios. Our method produces significantly more precise predictions with fewer candidate boxes compared to alternatives. The numbers indicate displayed boxes/total generated boxes, demonstrating our method's ability to \textit{reduce redundant detections while maintaining precision}. In the maritime scene (top), our approach avoids misclassifying background as objects, while in the transportation (middle) and indoor (bottom) examples, it generates tighter bounding boxes with minimal redundancy. These results validate our framework's effectiveness in remedying the Astigmatism problem.}
    \vspace{-2mm}
    \label{fig:bbox}
\end{figure}

\vspace{-1mm}
Figure~\ref{fig:bbox} demonstrates the effectiveness of our method in alleviating the Astigmatism problem via more focused and compact detections. Our approach consistently produces fewer and more accurate candidate boxes (e.g., reducing from 100 to 9 boxes in maritime scenes) while maintaining precise localization. Across domains, our model avoids false positives from background textures, generates tighter boxes under occlusion, and handles cluttered environments. These results visually confirm that our framework successfully transforms dispersed attention into object-centric focus, directly tackling the attention dispersion in CD-FSOD.
\vspace{-2mm}
\vspace{-2mm}

\subsection{Parameter Analysis and Discussion}
\subsubsection{Analysis of Background Text Quantity}

Table \ref{tab:text_ablation} reveals the impact of background text description quantity on model performance. 
Gains diminish beyond 200 descriptions—the largest increment occurs from 150 to 200 descriptions (+1.39\%), with diminishing returns thereafter (+0.08\% at 400 descriptions). Therefore, 200 descriptions provides the optimal performance-efficiency trade-off.
Notably, our complete framework introduces minimal \textbf{computational overhead} with negligible additional parameters and modest memory increases, demonstrating excellent deployment efficiency (detailed analysis in Appendix).

\vspace{-1mm}
\begin{table}[t]
\caption{Effect of background text quantity on performance. Eff. Ratio = $\Delta$AP / $\Delta$Mem., indicating AP gain per MB increase.}
\vspace{-2mm}
\centering
\scriptsize
\setlength{\tabcolsep}{5pt}
\begin{tabular}{c|cc|c|c}
\hline
\multirow{2}{*}{\textbf{Text Count}} & \multicolumn{2}{c|}{\textbf{Performance}} & \multirow{2}{*}{\textbf{GPU Mem. (MB)} $\downarrow$} & \multirow{2}{*}{\textbf{Eff. Ratio} $\uparrow$} \\
\cline{2-3}
 & \textbf{AP (\%)} $\uparrow$ & \textbf{Gain} $\uparrow$ & & \\
\hline
50 & 39.34 & - & 8762 & - \\
100 & 40.31 & +0.97\% & 8850 & 1.10\% \\
150 & 41.48 & +1.17\% & 8943 & 1.26\% \\
\textbf{200} & \textbf{42.87} & \textbf{+1.39\%} & \textbf{9027} & \textbf{1.65\%} \\
250 & 43.47 & +0.60\% & 9108 & 0.74\% \\
300 & 43.80 & +0.33\% & 9178 & 0.47\% \\
350 & 43.97 & +0.17\% & 9217 & 0.44\% \\
400 & 44.05 & +0.08\% & 9245 & 0.29\% \\
\hline
\end{tabular}
\label{tab:text_ablation}
\vspace{-6mm}
\end{table}

\subsubsection{Analysis of Background Alignment Loss Weight}

Figure~\ref{fig:parameter_analysis}(left) demonstrates the impact of TSA loss weight on our approach's effectiveness. The optimal performance occurs at $\lambda=10^3$, showing substantial improvement over baseline, which confirms that the TSA strengthens the distinctions between center and peripheral regions. The performance drop at extreme values validates our design principle that balanced cross-modal alignment is essential—neither insufficient nor excessive background emphasis yields optimal results. This supports our framework's ability to transform dispersed attention patterns into focused object-centric features through calibrated textual-visual feature alignment.

\vspace{-1mm}

\subsubsection{Analysis of Negative Feature Threshold}
\begin{figure}[t]
    \centering
    \includegraphics[width=\linewidth]{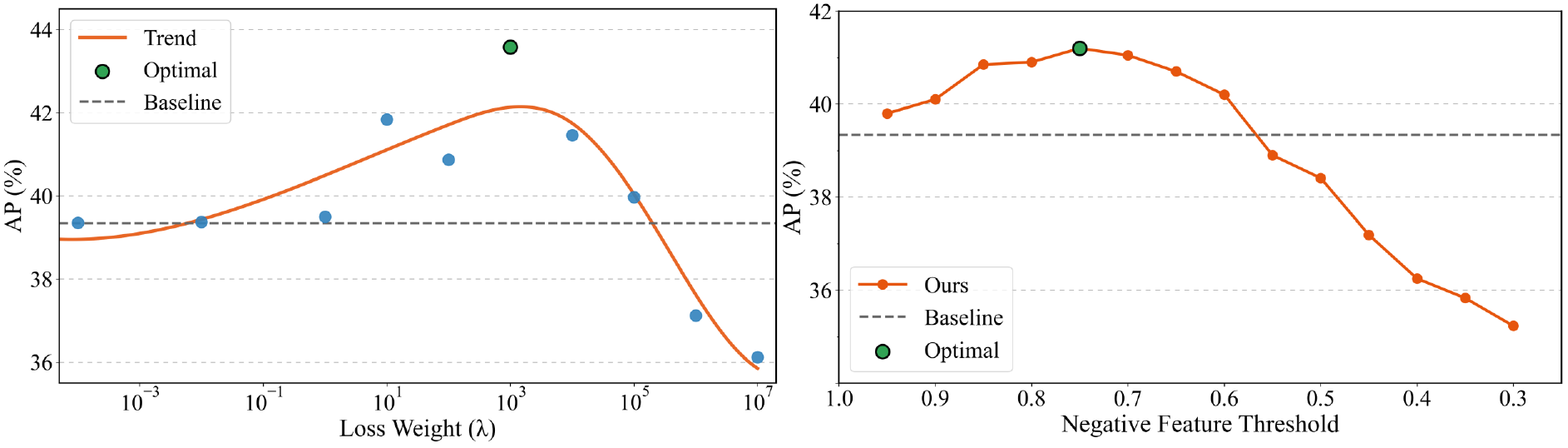}
    \vspace{-6mm}
    \caption{Analysis of key parameters in our approach. \textbf{Left}: Effect of TSA loss weight ($\lambda$) on detection performance. Performance remains stable when $\lambda \leq 1$, peaks at $\lambda = 10^3$, and declines beyond that, underscoring the importance of appropriate weighting in enhancing object-background separation. \textbf{Right}: Impact of threshold selection on NCM effectiveness. Performance remains above baseline across a wide threshold range (0.6-0.9), with optimal results at 0.75, demonstrating that selectively enhancing background features effectively reinforces object-background boundaries.}
    \label{fig:parameter_analysis}
    \vspace{-6mm}
\end{figure}

Figure~\ref{fig:parameter_analysis}(right) shows the effect of threshold selection in the Negative Context Modulation module. Our method consistently surpasses the baseline over a wide threshold range (0.6–0.9), with high thresholds still improving performance—indicating that updating a small portion of background features strengthens object–background separation. Performance peaks at 0.75, while lower thresholds risk degrading features due to excessive modification, confirming that targeted background enhancement effectively alleviates astigmatism without large-scale feature changes.
\vspace{-1mm}
\vspace{-1mm}
\subsubsection{Temperature and Weight Analysis in PPR}

\begin{figure}[htp]
    \centering
    \includegraphics[width=0.3\textwidth]{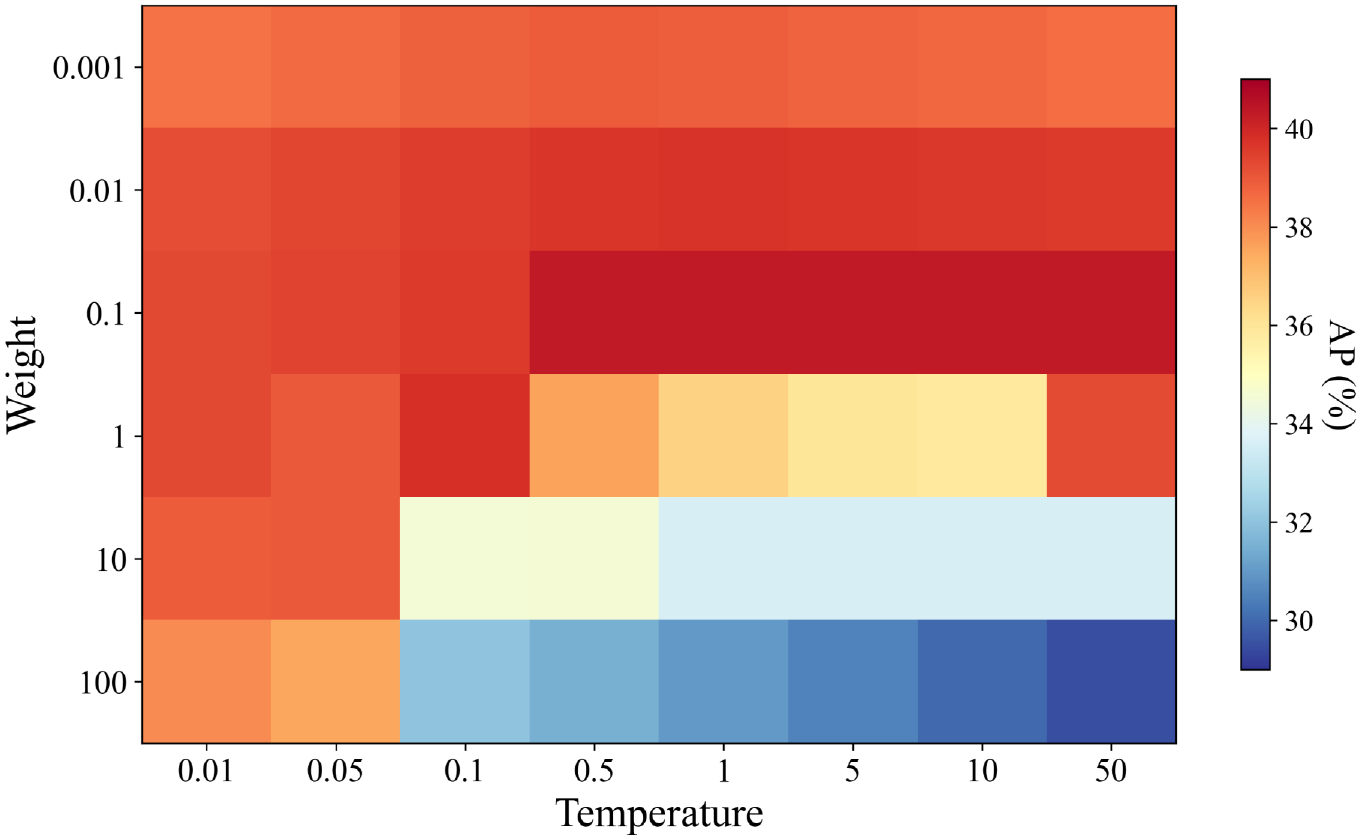}
    \vspace{-2mm}
    \caption{Parameter analysis of temperature-weight interactions in PPR. The heatmap shows peak performance at moderate weights (0.1-1.0) with robust temperature tolerance, while extreme weights ($\geq$1.0) cause significant performance degradation. This confirms that balanced prototype utilization is essential for effective attention refinement without disrupting original feature representations.}
    \label{fig:temperature_weight_heatmap}
\end{figure}

Figure~\ref{fig:temperature_weight_heatmap} shows the interaction between temperature and weight parameters in our PPR module. Moderate weights (0.1–1.0) consistently outperform extremes, with performance dropping sharply above 1.0. This supports our design principle that excessive foreground enhancement harms attention refinement. Robust performance across a wide temperature range within the optimal weight zone demonstrates the PPR module’s robustness across domains.\vspace{-2mm}


\vspace{-7mm}
\section{Conclusion}
\vspace{-2mm}
We identified the Astigmatism problem in CD-FSOD, where models exhibit dispersed attention in target domains, and proposed a comprehensive center-periphery attention refinement framework. Experiments demonstrate significant improvements over state-of-the-art methods.


\section*{Acknowledgments}
This work is supported by the National Natural Science Foundation of China under grants 62206102; the National Key Research and Development Program of China under grant 2024YFC3307900; the National Natural Science Foundation of China under grants 62436003, 62376103 and 62302184; Major Science and Technology Project of Hubei Province under grant 2025BAB011 and 2024BAA008; Hubei Science and Technology Talent Service Project under grant 2024DJC078; and Ant Group through CCF-Ant Research Fund. The computation is completed in the HPC Platform of Huazhong University of Science and Technology.

{
    \small
    \bibliographystyle{ieeenat_fullname}
    \bibliography{main}
}


\end{document}